\documentclass{article}

\usepackage{arxiv}

\usepackage[utf8]{inputenc} 
\usepackage[T1]{fontenc}    
\usepackage{hyperref}       
\usepackage{url}            
\usepackage{booktabs}       
\usepackage{amsfonts}       
\usepackage{amsmath,amssymb}
\usepackage{nicefrac}       
\usepackage{microtype}      
\usepackage{lipsum}
\usepackage{graphicx}
\usepackage[ruled,vlined]{algorithm2e}
\graphicspath{ {./images/} }
\usepackage{multirow}
\usepackage{array,booktabs}

\title{HYDRA - Hyper Dependency \\Representation Attentions
\thanks{*: The first two authors have equal contribution.}
}

\author{
 Nguyen Ha Thanh\\
  Japan Advanced Institute of Science and Technology
  \And
  Tran Duc Vu\\
  The Institute of Statistical Mathematics
  \And
  Dang Tran Binh \\
  Japan Advanced Institute of Science and Technology
  \And
  Bui Minh Quan\\
  Japan Advanced Institute of Science and Technology
  \And
  Nguyen Minh Phuong\\
  Japan Advanced Institute of Science and Technology
  \And
  Nguyen Le Minh\\
  Japan Advanced Institute of Science and Technology
   
}

\begin{document}
\maketitle
\begin{abstract}
Attention is all we need \textit{as long as we have enough data.}
Even so, it is sometimes not easy to determine how much data is enough while the models are becoming larger and larger.
In this paper, we propose HYDRA heads, lightweight pretrained linguistic self-attention heads to inject knowledge into transformer models without pretraining them again.
Our approach is a balanced paradigm between leaving the models to learn unsupervised and forcing them to conform to linguistic knowledge rigidly as suggested in previous studies.
Our experiment proves that the approach is not only the boost performance of the model but also lightweight and architecture friendly. 
We empirically verify our framework on benchmark datasets to show the contribution of linguistic knowledge to a transformer model. 
This is a promising result for a new approach to transferring knowledge from linguistic resources into transformer-based models.
\end{abstract}

\section{Introduction}
\label{sec:introduction}
Recently, pre-trained deep learning models have proven their effectiveness in a variety of tasks \cite{radford2018improving,devlin2018bert, brown2020language}.
One consequence of this is that the linguistic resources received less attention. 
In the digital age, the amount of data generated per second is enough to train a model capable of abstracting many different patterns.
With a sufficiently large amount of data, these models are shown to be able to abstract the linguistic features on their own \cite{liu2019linguistic}.
Even so, training the model unsupervised using data on the internet can generate noisy, biased models \cite{bender2021dangers} and make deep learning black boxes to humans.
Therefore, we want to find a balanced solution between those two extremes.


Pre-trained models derive their power from the attention mechanism especially multi-head self-attention \cite{vaswani2017attention}.
This mechanism mimics human observation, i.e. only cares about important aspects of the data that are relevant to the decision that needs to be made.
The model using attention mechanisms are often learned from data through unsupervised tasks. 
This helps to increase the performance of tasks without requiring much human intervention.
Even so, automation without explanation can carry risks when it is widely adopted.
Training models using only data can produce highly accurate but still fallacious models. Table \ref{tab:examples} shows a small dataset where if the model is based on the occurrence of the word "is", it can correctly classify 100\% of the sentiment labels of the samples.
Linguistically, "is" is a functional word and should not be used to make a decision in this case.
This is a contrived example, but the same thing can happen for a huge data set that humans are not capable of verifying \cite{chen2021evaluating}.
For that reason, before this work, there are proposals trying to force deep learning models to follow the knowledge of linguistic resources \cite{zhang2020semantics,zhou2019head,zhang2020sg}. 
We observe that such coercion could reduce the flexibility of the model and overall efficiency in exceptional cases.

\begin{table}
\caption{A hypothetical data in which a model can achieve 100\% accuracy without the need of understanding the semantic.}
\label{tab:examples}
\centering
\begin{tabular}{|l|l|}
\hline
\textbf{Sentence}                            & \textbf{Sentiment} \\ \hline
This \textbf{is} a great product.                     & Positive           \\ \hline
Awful service.                               & Negative           \\ \hline
This product \textbf{is} great.                       & Positive           \\ \hline
The battery of this product \textbf{is} very good.    & Positive           \\ \hline
I don't like this restaurant.                & Negative           \\ \hline
The song \textbf{is} perfect.                         & Positive           \\ \hline
This \textbf{is} another awesome product from Google. & Positive           \\ \hline
Nothing special.                             & Negative           \\ \hline
I think this product should not be sold.     & Negative           \\ \hline
It was a terrible experience.                & Negetive           \\ \hline
\end{tabular}
\end{table}

Finding a neutral solution, we investigate the possibility of injecting some linguistic knowledge for pretrained models to refer to without completely depending on them.
With this approach, the model can both use linguistic knowledge for the majority of cases and observe actual cases appearing in the data to adjust its understanding.
In addition, this approach guides the model to use the features that are accepted by humans, so that their decision-making is explainable by humans.
The challenge of this approach is to find a balance between the information learned in the pretraining and the injected knowledge.

We propose an unprecedented approach to incorporating linguistic knowledge into transformers architecture. 
This approach has 3 advantages compared to other methods.
First, this is a knowledge injection method that can take the edge of many different types of annotated resources in NLP. 
Second, the injected information is not rigid but only a reference source for the model to reach its final conclusion. 
Third, training and storing the appended component are extremely efficient, helping to solve environmental and financial problems.
Although we only experiment with dependency structure as the knowledge source in the scope of this paper, the approach is general to all linguistic knowledge which expresses relationships between internal elements of a sentence, is friendly with transformer architecture, and is extendable.
We believe this approach will create a new research direction in injecting knowledge into the transformer architecture.

Our paper includes the following main contributions as follows: 
\begin{itemize}
\item We first propose a novel framework to pretrain attention heads separately using the linguistic information without pretrain the whole model again. 
\item Next, we introduce HYDRA,  a knowledge-injected component to the transformer model, which is pretrained with dependency structure relations.
\item Finally, from the experimental result, we draw conclusions and future directions of this research.
\end{itemize}

In the remainder of this paper, we present related work in the attention mechanism and knowledge injection methods in Section~\ref{sec:background}, the detail of our method in Section ~\ref{sec:method}. Next, we investigate the effectiveness of the proposed method in comparison with the baseline in Section~\ref{sec:experiments}.  Finally, the conclusion and future works are presented in Section~\ref{sec:conclusions}.

\section{Related Work}
\label{sec:background}

\subsection{Attention Mechanism}
Transformer \cite{vaswani2017attention} has become the architecture that drastically changes the way deep learning is designed and used. With previous models such as LSTM or CNN, all the information of the inputs would be compressed into a single representation. 
The human need in NLP is to solve longer and longer sentences. With long inputs, these models run into problems, their representations are not able to store all the information of the input. 
Instead of storing a loop of information in the architecture like the LSTM, the transformer uses the attention mechanism to perceive constraints in the inputs. Attention can be described as aggregating a query and key-value pairs to the representation. 
The output (attention vector) is calculated as the weighted sum of the values, where the weight assigned to each value is computed by the query with the corresponding key. 
In the paper, the authors recommend of a combination of multiple aspects to the final representations, so-called multi-head attention. This mechanism allows the model to view data with different aspects to make decisions.

\subsection{Knowledge Injection}

Knowldge injection is the method of providing deep learning model with external knowledge.
In their paper, Nguyen and colleagues \cite{nguyen2021knowledge} use the similarity of the distribution as additional knowledge for the learning process. The knowledge was injected into the neural network (LSTM) to guide the model to make better predictions. The author also conduct the experiment to extract the knowledge from the data and use it to guide the model in another task. However, with the advent of transformers, LSTM gradually became obsolete.

Pretrained transformer-based language models such as BERT \cite{devlin2018bert} have demonstrated their effectiveness in many NLP tasks. 
However, these bulky models also reveal weaknesses through unsupervised learning of data \cite{chen2021evaluating,bender2021dangers}. 
As a result, using additional knowledge is a promising way to improve the models. 
Recent works \cite{zhang2020sgnet,sachan-etal-2021-syntax} discover that the performance of transformer models could be improved by incorporating either semantic information or syntactic information into the transformer models. 


To encode syntactic information into a transformer model, the authors \cite{zhang2020sgnet} consider using a syntactic representation to incorporate it into the attention mechanism so-called Syntax-guided network(SG-NET). The key component in SG-NET is the SDOI-SAN, which is a combination of syntactic dependency of interest (SDOI) and self-attention network (SAN). SG-NET used both SDOI-SAN and SAN of the original transformer model for better linguistics-inspired representation. 
This is a matrix $n \times n$ with $n$ as the number of words in the input sentence. An element in this metric is assigned a value of 1 if there is a dependency link of the words in the corresponding row and column, otherwise, it is assigned to 0. 

With the same idea of injecting knowledge to the model, Zhang et al. \cite{zhang2020semantics} proposes an architecture with combined semantic information and attention mechanism, so-called Semantics-guided neural network (SGN). The authors assign semantic labels for every token, encode them and integrate this kind of knowledge to the pretrained model using a convolutional neural network. We observe that in terms of implementation, this design is not friendly with the existing architecture of transformers and is hard to extend.

\section{Method}
\label{sec:method}
Observing the limitations of the previous methods, in this paper, we propose a novel partial pretraining method on the Transformer architecture.
Attention heads are essentially abstractions of data from different perspectives formed during the pretraining process.
The information in the heads interacts with each other to make the final decision.
So, we want to verify the possibility of pretraining the heads seperately with linguistic information and then incorporate them into the general architecture as an ordinary component.

\begin{figure*}
\centering
\includegraphics[width=0.8\textwidth]{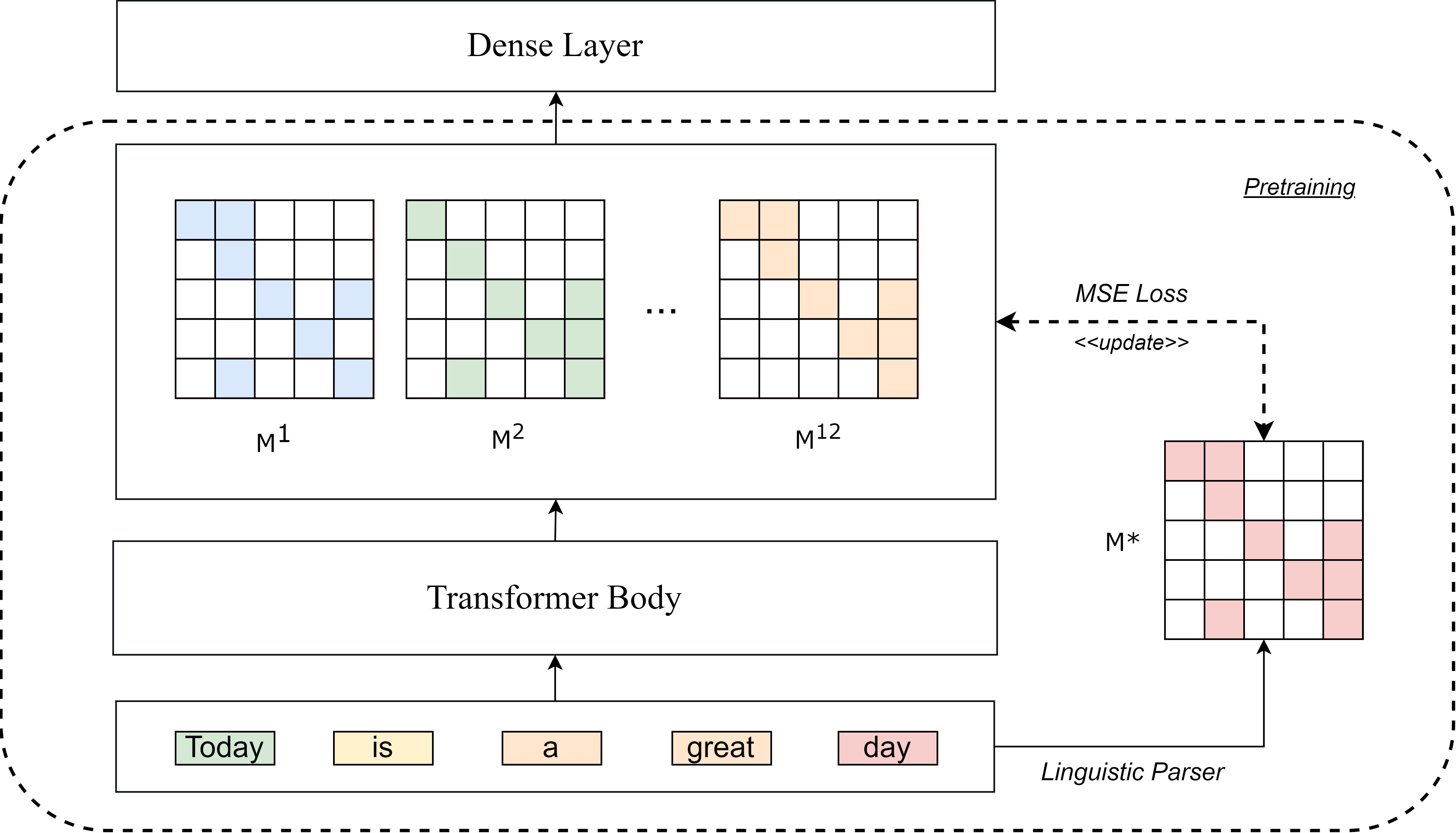}
  \caption{The general architecture of the framework. \textit{Transformer Body} is the pre-trained model (e.g. BERT). SDOI matrix $(M^*)$ is the relation matrix parsed from an external Linguistic Parser tool (e.g. Stanford NLP tool or spaCy).  \label{fig:framework}}
\end{figure*}

The overall architecture of the proposed framework is shown in Figure \ref{fig:framework}. Based on the transformer's architecture, we add one more transformer layer at the end of the architecture containing heads, which are pretrained with linguistic knowledge.
The model can choose to use the information of these heads or ignore it with the residual connections. 
Therefore, we can use linguistic knowledge as a reference source for the model without forcing it to accept such knowledge rigidly.
Since the idea is derived with a dependency structure, we give this architecture an inspiring name, \textit{HYDRA}, which stands for \textit{Hyper Dependency Representation Attentions}.

\subsection{Pretraining HYDRA Heads}


Let the input be $X = \{x_1, x_2, ..., x_n\}$ with $n$ be the sequence length, we obtain an $n \times n$ matrix $M^*$ containing the linguistic relationship of the words as introduced by Zhang et al.~\cite{zhang2020sgnet}.
After passing the input through the transformer layers, we get hidden state $H_{l}$ at the last layer with $l$ be the number of layers of the original transformer body.
We initiate and append  $l+1^{th}$ layer containing HYDRA heads and pass $H_{l}$ to this layer.
With $W_q$, $W_k$ are learnable parameters in the layer ${l+1}^{th}$, we calculate query and key vectors as in Equations \ref{eq:qvec} and \ref{eq:kvec}:
\begin{align}
q_{l+1}^h &= W_q \cdot H_{l}     \label{eq:qvec} \\
k_{l+1}^h &= W_k \cdot H_{l}    \label{eq:kvec}
\end{align}

\noindent With $d_k$ be the dimension of the key matrix $k_{l+1}^h$, we calculate the attention matrix for each head $M^h$ following Equation~\ref{eq:attn_mtx}:

\begin{equation}
\label{eq:attn_mtx}
M^h = \frac{q_{l+1}^h \cdot {k_{l+1}^h}^\intercal}{\sqrt{d_k}}
\end{equation}

\noindent We then minimize the element-wise MSE loss ($\mathcal{L}$) between $M^*$ and $M^h$ for every HYDRA head:
\begin{align}
\mathcal{L} &= \frac{1}{n^2} \sum_{i=0}^{n^2} (M^*_i - M^h_i)^2 
\end{align}
\noindent where $i$  $(0 \leq i < n^2 )$  is the index of element in flatten attention matrix SDOI ($M^*_i$). The weights of the transformer body are frozen and the weights in $Q_{l+1}^h$ and $K_{l+1}^h$ in every HYDRA heads are updated via the backpropagation process.

After pretraining, we get the HYDRA heads corresponding to the transformer body coupled to it. These HYDRA heads are saved in the storage and ready to be delivered as a form of deep learning resource. Compared to storing pretrained transformer models, storing HYDRA heads saves much more storage space. A transformer model checkpoint can be measured in gigabytes, while a HYDRA head checkpoint in our experiment is less than 10 megabytes.

\subsection{Fine-tuning with HYDRA Heads}
Different from approaches that use linguistic knowledge that is rigidly injected into the model, we attach pretrained HYDRA heads to the transformer bodies. These models can refer to but are not limited by the linguistic knowledge learned in the HYDRA heads.

At this phase, the weights of the transformer body and HYDRA head are both updated to match the training data. 
The final model now contains $l+1$ layers, first $l$ layers are from the original transformer body and the last layer contains HYDRA heads.
We simply pass the $H_{l+1}$ to a fully connected layer to get the logits for each downstream task. 
The loss function of each downstream task is defined similarly to the work \cite{devlin2018bert}. 
We have experiments to observe the performance of the model in Section \ref{sec:experiments}. Intuition in this phase is that by acquiring linguistic knowledge, the system can properly model the problem rather than try to fit into the local minima.

\section{Experiments}
\label{sec:experiments}

\subsection{Experimental Settings}
\subsubsection{Pretraining Stage}
In the pretrain phase, we collect data from Wikimedia Downloads\footnote{https://dumps.wikimedia.org/} using WikiExtractor package~\cite{Wikiextractor2015}. 
We use the variant of dependency parser described by Honnibal and Johnson~\cite{honnibal2015improved}, provided by spaCy\footnote{https://spacy.io/}.
After data processing, we obtain 330,000 samples for training and 50,000 samples for validation.

We use BERT~\cite{devlin2018bert} as the base model to pretrain the HYDRA heads. 
We set the maximum sequence length to 512 and only use sentences whose length does not exceed this number to pretrain the HYDRA heads. 
The goal of this setting is to help the model learn the linguistic structure of complete sentences of the longest possible length.
With 1.8 million parameters, in our experiment, it takes only one or two epochs for the HYDRA heads to reach optimal loss on both the training set and the validation set.

\subsubsection{Task Adaptation Stage}
\label{sec:task_adap}
To understand the model's behavior and the method's effectiveness, we run experiments and compare the performance of the HYDRA variants with the corresponding non-HYDRA baselines.
The experiments are run on datasets and standard settings of several famous benchmarks in natural language processing which include:

\begin{itemize}
    \item QNLI~\cite{rajpurkar2016squad}, MNLI~\cite{N18-1101,bowman2015large}, RTE~\cite{dagan2005pascal,bar2006second,giampiccolo2007third,bentivogli2009fifth} : Natural Language Inference and Texture Entailment
        \begin{itemize}
          \item Metric for QNLI and RTE: Accuracy.
          \item Metric for MNLI: Accuracy for both matched (m) and mismatched (mm) versions.
        \end{itemize}
    \item QQP~\cite{iyer2017qqp} and STS-B~\cite{cer2017semeval}: Semantically Equivalence Judgment
        \begin{itemize}
          \item Metric for QQP: Accuracy.
          \item Metric for STS-B: Pearson Spearman Correlation.
        \end{itemize}
    \item SQuAD~\cite{rajpurkar2016squad} and SQuAD 2.0~\cite{rajpurkar2018know}: Question Answering
        \begin{itemize}
          \item Metric for SQuAD and SQuAD 2.0: Exact Match and Macro-averaged F1 Score.
        \end{itemize}

\end{itemize}

\noindent With these benchmarks, we can investigate in detail the effectiveness of the proposed method. We conduct experiments with both the cased and uncased versions of BERT and average the performance.

\subsection{Experimental Result and Discussion}

Table \ref{tab:result} shows the results of the models on the benchmarks and metrics described in Section~\ref{sec:task_adap}. 
BERT is a very strong baseline, it achieved high performance on all benchmarks.
Even so, our model still can slightly improve the results of BERT with the appended HYDRA heads.
This result supports the hypothesis that linguistic knowledge pretrained in HYDRA heads can boost the performance of the vanilla model.

In addition to the performance improvement, this is a novel approach to inject linguistic knowledge into language models.
One notable feature of our improvement is that this new component is lightweight, requires low pretraining computation cost and storage.
With this paradigm, we can improve the bulky transformers models without pretraining the whole network again or force them to follow rigid linguistic rules.

\begin{table}
\centering
\caption{Experimental results on dev set of benchmark datasets\label{tab:result}}
\begin{tabular}{|l|c|c|}
\hline
\textbf{Benchmark} & \textbf{BERT} & \textbf{BERT HYDRA} \\ \hline
QNLI               & 0.9065                    & 0.9090                          \\ \hline
MNLI\_m            & 0.8373                    & 0.8401                          \\ \hline
MNLI\_mm           & 0.8419                    & 0.8458                          \\ \hline
RTE                & 0.6300                    & 0.6336                          \\ \hline
QQP                & 0.9062                    & 0.9067                          \\ \hline
STS-B              & 0.8800                    & 0.8812                          \\ \hline
SQuAD EM           & 0.8105                    & 0.8124                          \\ \hline
SQuAD F1           & 0.8838                    & 0.8851                          \\ \hline
SQuAD 2.0 EM       & 0.7137                    & 0.7161                          \\ \hline
SQuAD 2.0 F1       & 0.7458                    & 0.7480                          \\ \hline
\end{tabular}
\end{table}

\section{Conclusions}
\label{sec:conclusions}
This paper proposes an architecture-friendly and extensible method to improve the effectiveness of the transformer-based language models by pretraining and appending new knowledge-guided heads to their architecture. We conduct the experiment with BERT as the base model and the dependency information as the external knowledge. Our experiment shows that our lightweight component can help to boost the performance of transformer models and provide a flexible paradigm to partially inject the knowledge into bulky models without pretraining them again. Extending this work, we can analyze the possibility of pretraining this component with different knowledge forms for problems in narrower domains or explore the potential of this approach for other data such as photos or~videos.

\bibliographystyle{plain}      
\bibliography{./references}   

\end{document}